\documentclass{article} 
\usepackage{iclr2026_conference,times}

\usepackage{amsmath,amsfonts,bm}

\def\eqref#1{equation~\ref{#1}}

\def\1{\bm{1}}

\DeclareMathAlphabet{\mathsfit}{\encodingdefault}{\sfdefault}{m}{sl}
\SetMathAlphabet{\mathsfit}{bold}{\encodingdefault}{\sfdefault}{bx}{n}

\usepackage{hyperref}
\usepackage{url}
\usepackage{booktabs}
\usepackage{graphicx}
\graphicspath{{../figs/}{figs/}}
\usepackage{multirow}
\title{Recovering Cloud Microstructures with Cascaded Diffusion Inversion}

\author{
  Hanan Gani\textsuperscript{1,3},
  Guy Pulik\textsuperscript{2},
  Daniel Rosenfeld\textsuperscript{2},
  Duncan Watson-Parris\textsuperscript{3},
  Salman Khan\textsuperscript{1} \\
  \textsuperscript{1}Mohamed Bin Zayed University of Artificial Intelligence \\
  \textsuperscript{2}Hebrew University of Jerusalem \\
  \textsuperscript{3}University of California, San Diego \\
  Correspondence: hanan.ghani@mbzuai.ac.ae, salman.khan@mbzuai.ac.ae
}
\iclrfinalcopy 
\begin{document}

\maketitle
\begin{abstract}
High-resolution satellite imagery is critical for observing fine-scale cloud structures that inform weather modification strategies like cloud seeding for rain-enhancement. However, the spatial resolution of current geostationary and polar-orbiting satellites is often insufficient for capturing small cloud features. Current super-resolution methodologies are suited for natural images and, therefore, struggle to generalize to satellite-captured spectral images of cloud cover. To address this, we propose a two-stage diffusion-based super-resolution framework to enhance the resolution of multi-spectral cloud microstructures by a factor of $4\times$. Specifically, we use inverse diffusion to recover the high resolution properties from low resolution. Stage 1 utilizes real-world paired data to learn robust degradation handling and inter-sensor alignment, while Stage 2 employs a self-supervised internal downgrading of high resolution data to refine structural learning and texture synthesis. Our approach outperforms the state-of-the-art transformer and diffusion-based baselines in both reconstruction accuracy and visual quality. We demonstrate that the two-stage method better captures fine cloud microstructures (e.g. convective turrets and cloud gaps) that are crucial for effective cloud seeding decisions. Ablation studies confirm the complementary benefits of the two stages: Stage 1 excels in coarse structural fidelity, while Stage 2 contributes enhanced detail and realism. These results highlight a practical path toward improving cloud microphysics analysis and as a step towards utilizing AI for climate and sustainability. Our code and models are publicly available at: \color{magenta}\url{ https://github.com/hananshafi/superresolution-cloud-microphysics}
\end{abstract}

\section{Introduction}
\label{sec:intro}
High-resolution meteorological satellite cloud imagery is essential for diagnosing localized weather phenomena and for operations such as weather modification (e.g., cloud seeding), which relies on identifying suitable microphysical targets such as supercooled liquid regions \citep{wmo_weather_modification,noaa_cloudseeding}.
However, spatial and temporal resolution trade off sharply across current observing systems. Modern geostationary (GEO) imagers provide rapid refresh but remain spatially coarse: for example, GOES-R ABI offers $\sim$0.5--2\,km sampling depending on band \citep{noaa_abi}, while Meteosat Second Generation (MSG) provides full-disk imagery every 15 minutes from geostationary orbit \citep{eumetsat_msg}, and its $\textsc{seviri}$ imager samples at $\sim$3\,km at the sub-satellite point \citep{wmo_oscar_seviri}. Moreover, the effective GEO pixel footprint grows away from the sub-satellite point \citep{ceda_seviri_frp}, blurring sub-kilometer structures that are often diagnostically important.

Low-Earth-orbit (LEO) instruments such as $\textsc{viirs}$ resolve substantially finer detail (375\,m and 750\,m native resolutions at nadir) but provide limited temporal sampling at a fixed location, since global coverage is achieved via orbital revisits rather than continuous staring \citep{podaac_suomi_npp}. For cloud-focused applications, this motivates computational super-resolution (SR) that aims to recover LEO-like spatial detail while retaining GEO-like update frequency.

This setting is particularly challenging because the $\textsc{seviri}\rightarrow\textsc{viirs}$ (and $\textsc{msg}\rightarrow\textsc{mtg}$) mapping is cross-sensor and time-shifted, with imperfect alignment and non-stationary degradations. Deterministic SR methods optimized for pixel fidelity, including transformer-based models \citep{liang2021swinir}, tend to oversmooth high-variance cloud textures and degrade under cross-sensor distribution shift. Diffusion-based~\citep{song2021ddim,ho2020ddpm} SR can synthesize sharper detail, but may introduce artifacts or inconsistent structure when applied outside the natural-image priors they inherit \citep{Wang_2024_IJCV,Wang_2024_CVPR}.

\begin{figure}[t]
    \centering
    \includegraphics[width=0.9\linewidth]{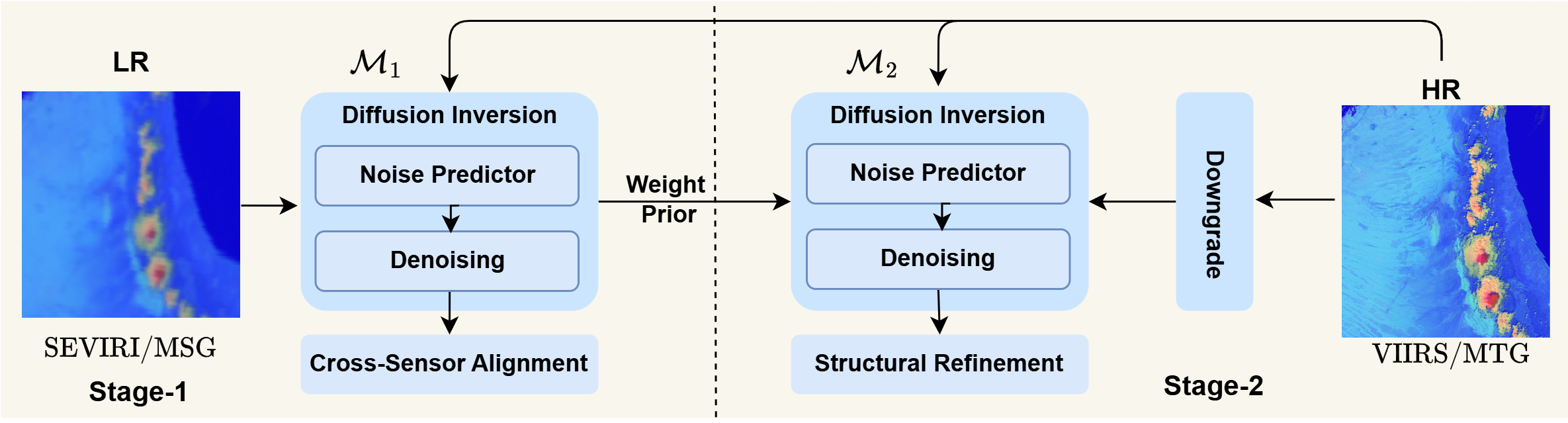}
    \caption{\textbf{Two-stage InvSR training.} Stage~1 trains InvSR on paired \textsc{seviri}$\rightarrow$\textsc{viirs}/\textsc{msg}$\rightarrow$\textsc{mtg} data to learn cross-sensor alignment. Stage~2 trains the same InvSR formulation on HR-only data with synthetic degradations to emphasize high-frequency cloud structures.}
    \label{fig:method_overview}
    \vspace{-1em}
\end{figure}

We address these limitations with a two-stage curriculum for diffusion inversion SR \citep{Yue_2025_CVPR} that explicitly decouples cross-sensor robustness from texture recovery (see Fig. \ref{fig:method_overview}). Stage~1 trains on real paired $\textsc{seviri}\rightarrow\textsc{viirs}$ / $\textsc{msg}\rightarrow\textsc{mtg}$ data to learn a physically consistent mapping under temporal and geometric mismatch. Stage~2 initializes from the Stage~1 prior and trains on HR-only imagery with synthetic degradations to learn cloud-specific fine-scale structure in a perfectly aligned setting. The resulting cascade improves both distortion and structure preservation, producing sharper yet more consistent reconstructions for multi-spectral cloud imagery.


\section{Related Work}
\label{sec:related_work}
\textbf{Image Super-Resolution.}
Deterministic SR methods, from bicubic interpolation and early CNN baselines~\citep{dong2016srcnn,kim2016vdsr} to modern transformer architectures~\citep{liang2021swinir,conde2022swin2sr}, typically minimize pixel-wise distortion, which often yields oversmoothed textures under large scaling factors and imperfect supervision. Diffusion-based~\citep{ho2020denoising,song2020denoising} SR~\citep{wang2023stablesr,Wang_2024_CVPR} can synthesize sharper detail by leveraging strong generative priors, but their stochastic generation can introduce artifacts or structurally inconsistent textures when transferred to new domains. We build on diffusion inversion~\citep{Yue_2025_CVPR}, which frames SR through an invertible latent-variable formulation. This perspective explicitly models lost high-frequency information and enables a curriculum that separates cross-sensor robustness from detail synthesis, improving structural fidelity without relying on unconstrained hallucination.\\
\textbf{Multi-Spectral Cloud Microphysics.}
Multi-spectral satellite observations are widely used to study cloud microphysics (e.g., optical thickness and effective radius). Geostationary sensors such as \textsc{SEVIRI} on MSG provide high temporal sampling for tracking cloud evolution but at coarse spatial resolution~\citep{schmetz2002seviri}. In contrast, polar-orbiting instruments such as \textsc{VIIRS} provide finer spatial detail but with limited revisit frequency~\citep{cao2013viirs}. Bridging this spatio-temporal gap is important for cloud-focused applications, motivating cross-sensor super-resolution. Our work targets this setting and additionally considers next-generation Meteosat Third Generation satellite system (MTG)~\citep{holmlund2021mtg}.

\section{Methodology}
\label{sec:methodology}

We formulate cloud super-resolution as an inverse problem: given a low-resolution (LR) observation $\mathbf{x}\in\mathbb{R}^{h\times w\times C}$ (\textsc{seviri}/\textsc{msg}), we recover a high-resolution (HR) microphysical field $\mathbf{y}\in\mathbb{R}^{H\times W\times C}$ (\textsc{viirs}/\textsc{mtg}), with $H\gg h$. The observation process is modeled as
\begin{equation}
\mathbf{x} = \mathcal{D}(\mathbf{y}) + \boldsymbol{\eta},
\label{eq:forward_model}
\end{equation}
where $\mathcal{D}$ subsumes blur, resampling, spectral mismatch, and temporal/geometric misalignment, and $\boldsymbol{\eta}$ denotes sensor noise. Recovering $\mathbf{y}$ from $\mathbf{x}$ is therefore ill-posed because multiple HR cloud structures can map to the same LR observation.

Following diffusion inversion~\citep{Yue_2025_CVPR}, we model the missing high-frequency content with a latent variable at diffusion step $t$ and learn a conditional inverse map
\begin{equation}
\hat{\mathbf{y}}_t = \mathcal{R}_{\theta}(\mathbf{x}, t) = \Phi^{-1}_{\theta}\!\big(\mathbf{x}, g_{\theta}(\mathbf{x}, t)\big),
\label{eq:inversion_map}
\end{equation}
where $g_{\theta}$ predicts the deterministic starting point of the reverse process from the LR input, and $\Phi^{-1}_{\theta}$ denotes the corresponding inversion-and-sampling operator. This lets the model remain anchored to the observed satellite signal while still recovering plausible fine-scale detail. We instantiate this idea as a \textit{cascaded invertible framework} that separates: (1) cross-sensor domain alignment and (2) high-frequency structure restoration.

\textbf{Stage 1: Cross-Sensor Alignment (Learning the Prior). }The primary challenge in satellite SR is the domain gap caused by sensor differences and spatiotemporal misalignment between geostationary and polar imagery. We train the first network, $\mathcal{M}_1$, on paired real-world data $\mathcal{D}_{\text{real}} = \{(\mathbf{x}_{\text{LR}}, \mathbf{y}_{\text{HR}})\}$. For each pair, the model reconstructs
\begin{equation}
\hat{\mathbf{y}}^{(1)}_t = \mathcal{R}_{\theta_1}(\mathbf{x}_{\text{LR}}, t),
\end{equation}
and optimizes
\begin{equation}
\mathcal{L}_{\text{S1}} =
\mathbb{E}_{(\mathbf{x},\mathbf{y})\sim\mathcal{D}_{\text{real}},\, t}
\left[
\|\hat{\mathbf{y}}^{(1)}_t - \mathbf{y}\|_1
\;+\;
\lambda_{\mathrm{perc}}\,\mathcal{L}_{\mathrm{perc}}(\hat{\mathbf{y}}^{(1)}_t,\mathbf{y})
\right].
\label{eq:stage1_loss}
\end{equation}
Because supervision comes from real pairs, Stage~1 primarily learns radiometric consistency, coarse geometry, and robustness to residual temporal/geometric mismatch.\\
\textbf{Stage 2: Structural Refinement. }
While Stage~1 aligns the data, the inherent noise in real-world pairings prevents it from learning sharp high-frequency textures. Stage~2, $\mathcal{M}_2$, focuses purely on recovering convective structures from aligned supervision. We construct a synthetic dataset $\mathcal{D}_{\text{syn}}$ by applying a degradation kernel $k$ to HR images,
\begin{equation}
\mathbf{x}_{\text{syn}} = \mathcal{D}_k(\mathbf{y}_{\text{HR}}),
\label{eq:synth_degrade}
\end{equation}
and optimize a second inverse model
\begin{equation}
\hat{\mathbf{y}}^{(2)}_t = \mathcal{R}_{\theta_2}(\mathbf{x}_{\text{syn}}, t),
\end{equation}
with loss
\begin{equation}
\mathcal{L}_{\text{S2}} =
\mathbb{E}_{\mathbf{y}\sim\mathcal{D}_{\text{HR}},\, t}
\left[
\|\hat{\mathbf{y}}^{(2)}_t - \mathbf{y}\|_1
\;+\;
\lambda_{\mathrm{perc}}\,\mathcal{L}_{\mathrm{perc}}(\hat{\mathbf{y}}^{(2)}_t,\mathbf{y})
\right].
\label{eq:stage2_loss}
\end{equation}
Because $\mathbf{x}_{\text{syn}}$ and $\mathbf{y}_{\text{HR}}$ are perfectly aligned, Stage~2 concentrates on recovering thin cloud filaments, convective cores, and local gradient structure that are difficult to learn from noisy cross-sensor pairs alone.\\
\textbf{Inference.}
At test time, given a real LR input $\mathbf{x}_{\text{LR}}$, we first compute a cross-sensor-consistent base estimate $\hat{\mathbf{y}}_{\text{base}} = \mathcal{R}_{\theta_1}(\mathbf{x}_{\text{LR}}, t^\star)$ using a short reverse chain (typically 1-step for efficiency). When additional refinement is beneficial, we apply the cascaded pass $\mathbf{x}_{\text{LR}}\xrightarrow{\mathcal{M}_1}\hat{\mathbf{y}}_{\text{base}}\xrightarrow{\mathcal{M}_2}\hat{\mathbf{y}}_{\text{final}}$, where the second stage sharpens structures learned under aligned synthetic supervision.

\section{Experimentation}
We base our approach on diffusion inversion SR~\citep{Yue_2025_CVPR}. The data for training and validation multi-spectral cloud cover is sourced from the UAE Rain Enhancement Program (UAEREP) under the National Center for Meteorology. Our data includes 9k train and 1k validation low- and high-resolution pairs corresponding to \textsc{seviri} and \textsc{viirs}, respectively. We further scale our approach to channel-wise data obtained across \textsc{msg} and \textsc{mtg} satellites, for which we obtained around 2250 training pairs and 500 validation pairs. Training both stages takes around 5 days on two Nvidia A100 40GB GPUs. We compare against an existing transformer-based approach~\citep{liang2021swinir} and two recent diffusion-based approaches~\citep{wang2023stablesr,Wang_2024_CVPR}. For empirical evaluation, we use PSNR, gradient preservation, and perceptual distance ratio to assess super-resolution quality. Refer to the appendix for additional details on data specifications and channel alignment.

\begin{table*}[t]
\caption{Quantitative SR results on $\textsc{seviri}\rightarrow\textsc{viirs}$ and $\textsc{msg}\rightarrow\textsc{mtg}$.}
\label{tab:results_main}
\centering
\resizebox{0.9\textwidth}{!}{
\begin{tabular}{lccc|ccc}
\toprule
& \multicolumn{3}{c|}{\textbf{\textsc{seviri}$\rightarrow$\textsc{viirs}}} &
\multicolumn{3}{c}{\textbf{\textsc{msg}$\rightarrow$\textsc{mtg}}} \\
\cmidrule(lr){2-4}\cmidrule(lr){5-7}
Model &
PSNR$\uparrow$  & Grad.$(\text{ideal=1.0})$ & Percep$\downarrow$ &
PSNR$\uparrow$  & Grad.$(\text{ideal=1.0})$ & Percep$\downarrow$ \\
\midrule
SwinIR      &18.37   &0.19  &0.45  &15.91  &0.53  &0.75    \\
StableSR   &20.52   &1.72  &0.44  &18.71  &2.94  & 0.42   \\
SinSR     &20.69   &0.46  &0.37  &24.0  & 0.96 & 0.30   \\
\midrule
\textbf{Ours} &\textbf{21.25}   &\textbf{1.06}  & \textbf{0.28} & \textbf{24.0}  &\textbf{1.03}  &\textbf{0.29}    \\
\bottomrule
\end{tabular}}
\label{tab:results}
\vspace{-1em}
\end{table*}

\begin{figure}[t]
    \centering
    \includegraphics[width=0.9\linewidth]{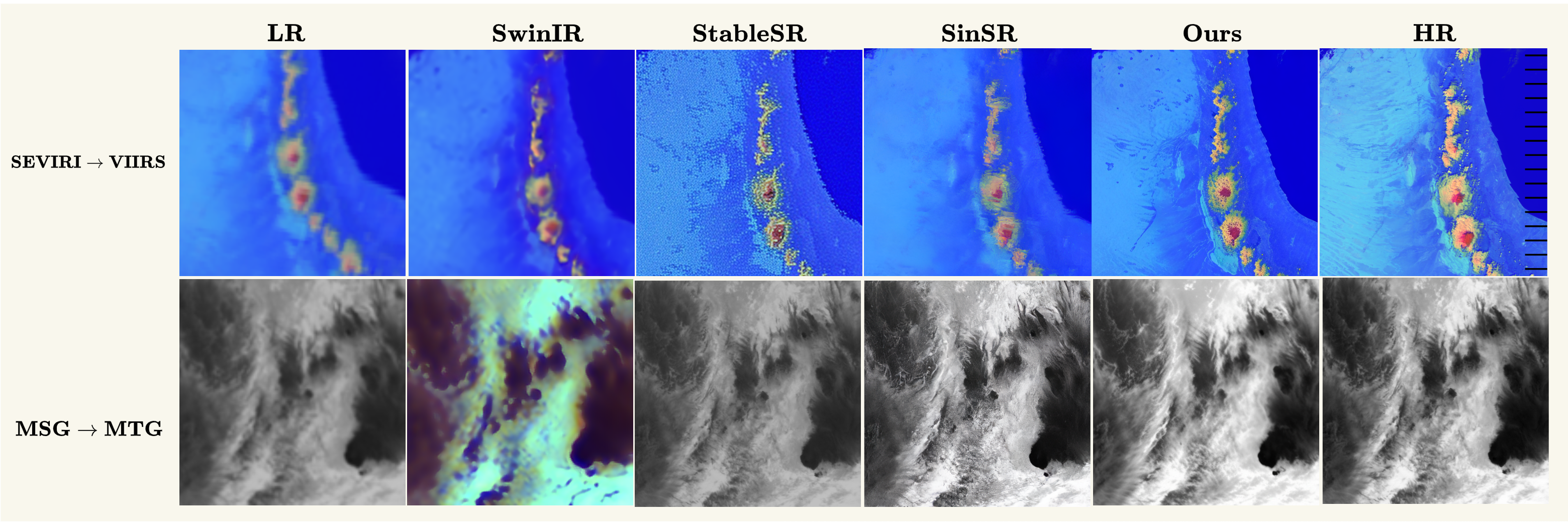}
    \caption{\textbf{Qualitative Comparison.} Most baselines produce blurry and over-sharpened outputs, however, our approach ticks the correct balance between sharpness and smoothing, constantly adhering to the ground-truth HR structure (best viewed in zoom).}
    \vspace{-3mm}
    \label{fig:qualitative}

\end{figure}
\textbf{Quantitative Results. }Table \ref{tab:results} summarizes performance on \textsc{seviri}$\rightarrow$\textsc{viirs} pairs and channel-wise \textsc{msg}$\rightarrow$\textsc{mtg} pairs using three complementary metrics: PSNR (distortion), gradient preservation ratio (structure; ideal = 1), and perceptual distance (lower is better). On \textsc{seviri}$\rightarrow$\textsc{viirs}, our method achieves the best overall balance, improving PSNR to 21.25 dB while producing the lowest perceptual distance (0.28). In contrast, SwinIR underperforms on cross-sensor data (18.37 dB) and severely under-recovers high-frequency structure (grad. ratio 0.19). StableSR yields a high gradient ratio (1.72), indicating over-sharpening relative to the target gradient statistics with high perceptual distance (0.44) while taking 100 denoising steps. SinSR improves over SwinIR in both PSNR and perceptual distance, but remains structurally under-sharp (gradient ratio 0.46) compared to our near-ideal 1.06.
On \textsc{msg}$\rightarrow$\textsc{mtg}, our method matches the best PSNR (24.0 dB) while achieving the lowest perceptual distance (0.29) and the closest-to-ideal gradient ratio (1.03). StableSR again exhibits a strongly inflated gradient ratio (2.94), consistent with oversharpening/artifacts, whereas SwinIR lags substantially in both PSNR (15.91 dB) and perceptual distance (0.75). Overall, the results support our proposed two-stage diffusion-inversion training which improves fidelity (PSNR) and preserves physically meaningful structure (grad. ratio$\sim$1) without increasing perceptual error.
\begin{figure}[t]
    \centering
    \includegraphics[width=0.42\linewidth]{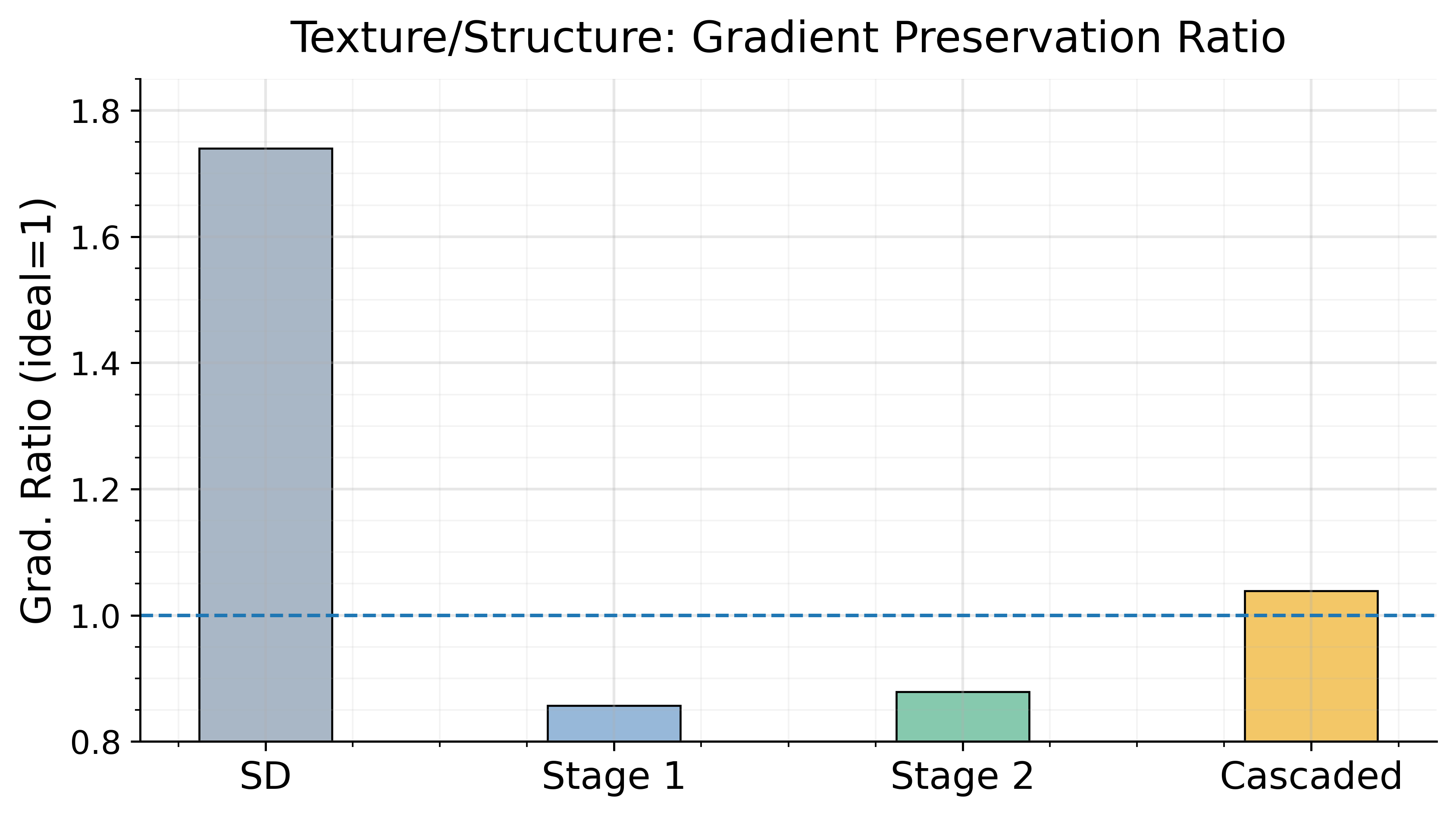}
    \caption{Effectiveness of each stage.}
    \label{fig:ablation}
    \vspace{-0.8\baselineskip}
\end{figure}

\textbf{Qualitative Analysis. }Fig. \ref{fig:qualitative} shows visual comparisons for \textsc{seviri}$\rightarrow$\textsc{viirs} (top) and \textsc{msg}$\rightarrow$\textsc{mtg} (bottom). On \textsc{seviri}$\rightarrow$\textsc{viirs}, SwinIR largely preserves the low-frequency appearance but over-smooths the narrow cloud band and convective cores, failing to recover the fine filaments visible in HR. StableSR produces noticeably sharper outputs, but the sharpening comes with widespread speckle / grid-like artifacts and noisy textures (inflated gradient ratio), which do not correspond to the HR structures. SinSR recovers some mid-frequency details but still misses thin streaks and tends to smooth background texture. In contrast, our method reconstructs coherent high-frequency structure (crisper boundaries and more faithful cloud-cell shapes) while avoiding spurious texture, yielding the closest visual match to HR.
On \textsc{msg}$\rightarrow$\textsc{mtg}, SwinIR exhibits severe appearance/contrast artifacts under the cross-sensor setting, while StableSR and SinSR sharpen cloud regions but can introduce granularity or wash out fine wisps. Our approach preserves the mesoscale cloud streaks and local gradients more consistently, producing fine-scale patterns that align with the HR target without obvious hallucinations.

\textbf{Discussion. }Most SR baselines implicitly assume (a) a known, stationary degradation (often bicubic), (b) tightly aligned LR/HR supervision, and (c) a single-sensor imaging model. In $\textsc{seviri}\rightarrow\textsc{viirs}$ and $\textsc{msg}\rightarrow\textsc{mtg}$, these assumptions are violated: the mapping is cross-sensor, time-shifted, and affected by different PSFs, noise characteristics, and spectral responses. As a result, distortion-driven models tend to blur (regression-to-the-mean under misalignment), while perceptual/diffusion baselines can generate sharp but unreliable textures.
Fig. \ref{fig:ablation} helps disentangle the role of each training stage using the gradient preservation ratio (ideal = 1.0). The single-stage variants remain under-sharp: Stage-1 improves robustness but retains smoothing (ratio 0.86), and Stage-2 slightly improves structure (ratio 0.88) but is still conservative. The cascaded two-stage training moves the ratio close to the physically desirable regime (1.04), indicating substantially better recovery of fine-scale cloud boundaries and filaments. In contrast, the diffusion baseline (SD) overshoots the target gradients (ratio 1.74), consistent with over-sharpening artifacts rather than faithful structure.
Overall, the plot supports our main claim: Stage-1 provides cross-sensor robustness, Stage-2 injects aligned structural supervision, and their combination is necessary to achieve near-ideal performance.
\vspace{-2em}
\section{Conclusion}
We presented a two-stage diffusion-inversion super-resolution framework for multi-spectral cloud imagery in challenging cross-sensor settings, including \textsc{seviri}$\rightarrow$\textsc{viirs} and \textsc{msg}$\rightarrow$\textsc{mtg}. The key idea is to decouple \emph{robust cross-sensor mapping} from \emph{fine-detail recovery}: Stage~1 leverages real paired observations to handle temporal and geometric mismatch, while Stage~2 uses HR-only self-supervision with synthetic degradations to recover high-frequency cloud structure under aligned training. Across both benchmarks, our approach achieves a stronger trade-off between distortion, structural fidelity (gradient preservation), and perceptual distance than transformer- and diffusion-based baselines, producing sharper yet more consistent reconstructions.

\clearpage
\bibliography{iclr2026_conference}
\bibliographystyle{iclr2026_conference}

\newpage
\appendix
\section{Appendix}
\subsection{Band Selection and Pair Construction}
\label{app:bands_alignment}

\subsubsection{Channels used and cross-sensor correspondence}
A central challenge in cross-sensor super-resolution is that ``matching channels'' are only approximate: sensors differ in their spectral response functions (SRFs), viewing geometry, calibration, point-spread functions (PSFs), and noise characteristics. Consequently, even when two bands are nominally similar (e.g., ``0.6$\mu$m reflectance''), the measured radiance/reflectance/brightness temperature will not be identical. We therefore choose channels that (i) measure the \emph{same physical quantity} (reflectance or brightness temperature), and (ii) are the \emph{closest available spectral counterparts} across sensors. Any residual mismatch is treated as part of the domain gap handled by Stage~1 real-pair training.

\paragraph{\textsc{SEVIRI}$\rightarrow$\textsc{VIIRS} (composite).}
For \textsc{seviri}$\rightarrow$\textsc{viirs} we represent each scene using a 3-channel ``day microphysics'' style composite consisting of two reflective bands and one thermal band. This choice provides a compact but informative representation of cloud morphology and thermodynamic structure: the reflective channels emphasize cloud optical properties and fine boundaries, while the thermal channel captures cloud-top temperature patterns and mesoscale organization.

\begin{itemize}
    \item \textbf{\textsc{viirs}.} We use I-band SDRs: \emph{I\_01} (reflectance), \emph{I\_03} (reflectance), and \emph{I\_05} (brightness temperature).
    \item \textbf{\textsc{seviri}.} We use the closest \textsc{seviri} counterparts: \emph{VIS006} (0.6$\mu$m reflectance), \emph{IR\_016} (1.6$\mu$m reflectance), and \emph{IR\_108} (10.8$\mu$m brightness temperature).
\end{itemize}

In our preprocessed SEVIRI tensor (25 layers), longitude/latitude are stored at indices $[0,1]$, and the three composite channels used for learning correspond to indices $[6,8,22]$ (i.e., \texttt{-3} for the thermal band), consistent with the VIS006/IR\_016/IR\_108 triplet. We apply fixed clipping and normalization to stabilize training and enable consistent dynamic ranges across sensors: reflective channels are scaled to $[0,1]$ and $[0,0.8]$, and the thermal channel is clipped to $[203,323]$\,K prior to normalization.

\begin{table}[ht]
\centering
\small
\setlength{\tabcolsep}{6pt}
\begin{tabular}{lll}
\toprule
\textbf{Quantity} & \textbf{SEVIRI (MSG)} & \textbf{VIIRS} \\
\midrule
Visible reflectance & VIS006 (0.6$\mu$m) & I\_01 (0.64$\mu$m) \\
NIR/SWIR reflectance & IR\_016 (1.6$\mu$m) & I\_03 (1.61$\mu$m) \\
Thermal IR (BT) & IR\_108 (10.8$\mu$m) & I\_05 (11.45$\mu$m) \\
\bottomrule
\end{tabular}
\caption{Channel correspondence used to form the SEVIRI$\rightarrow$VIIRS 3-channel composites. Channels are matched by physical quantity and nearest spectral counterpart. Differences in SRF/PSF remain and constitute part of the cross-sensor gap addressed in Stage~1.}
\label{tab:seviri_viirs_channels}
\end{table}

\paragraph{\textsc{MSG}$\rightarrow$\textsc{MTG} (channel-wise pairs).}
For \textsc{msg}$\rightarrow$\textsc{mtg} we create \emph{channel-wise} paired samples to evaluate and train across a broader set of spectral bands. We load \textsc{mtg}/\textsc{fci} L1C using Satpy (\texttt{reader=fci\_l1c\_nc}) and \textsc{msg}/\textsc{seviri} L1B using Satpy (\texttt{reader=seviri\_l1b\_native}), then match each \textsc{mtg} band to its closest \textsc{msg} counterpart by spectral region and physical quantity (reflectance vs brightness temperature). We restrict to the overlapping subset of channels for which a clear correspondence exists.

\begin{table}[t]
\centering
\small
\setlength{\tabcolsep}{6pt}
\begin{tabular}{lll}
\toprule
\textbf{Quantity} & \textbf{MTG/FCI} & \textbf{MSG/SEVIRI} \\
\midrule
Visible reflectance & vis\_06 & VIS006 \\
Visible reflectance & vis\_08 & VIS008 \\
NIR reflectance & nir\_16 & IR\_016 \\
Water vapor (BT) & wv\_63 & WV\_062 \\
Water vapor (BT) & wv\_73 & WV\_073 \\
Thermal IR (BT) & ir\_38 & IR\_039 \\
Thermal IR (BT) & ir\_87 & IR\_087 \\
Thermal IR (BT) & ir\_97 & IR\_097 \\
Thermal IR (BT) & ir\_105 & IR\_108 \\
Thermal IR (BT) & ir\_123 & IR\_120 \\
Thermal IR (BT) & ir\_133 & IR\_134 \\
\bottomrule
\end{tabular}
\caption{Channel-wise correspondence used for MSG$\rightarrow$MTG. Channels are matched by physical quantity and nearest spectral region; exact SRFs differ across sensors.}
\label{tab:msg_mtg_channels}
\end{table}

Reflectance channels are scaled to $[0,1]$ (including percent-to-fraction conversion when needed). Brightness temperature channels are normalized using a fixed physical range of $[180,330]$\,K before saving patches (16-bit PNG), ensuring consistent scaling across channels.

\subsubsection{Grid alignment for paired samples}
\paragraph{SEVIRI--VIIRS alignment (geolocation-based regridding).}
To create aligned \textsc{seviri}--\textsc{viirs} pairs we use per-pixel geolocation grids. \textsc{viirs} longitude/latitude are read from the \textsc{viirs} geolocation product, while \textsc{seviri} longitude/latitude are read from the \textsc{seviri} grid (stored as the first two layers in the \textsc{seviri} tensor). We first identify a \textsc{seviri} crop whose latitude/longitude extent lies within the \textsc{seviri} scene bounds (via crop-wise min/max checks) to guarantee overlap. We then reproject \textsc{viirs} onto the \textsc{seviri} grid using nearest-neighbor interpolation in $(\mathrm{lon},\mathrm{lat})$ space (via \texttt{scipy.interpolate.griddata}), producing co-registered pairs on a shared grid for training and evaluation. Remaining pixel-level discrepancies can persist due to time offset, differing view angles, and residual navigation errors; these are handled implicitly by Stage-1 training on real paired data.

\paragraph{MSG--MTG alignment (Satpy area resampling).}
For \textsc{msg}--\textsc{mtg} we rely on Satpy's projection-aware resampling. We load \textsc{mtg}/\textsc{fci} L1C using Satpy (\texttt{reader=fci\_l1c\_nc}) and \textsc{msg}/\textsc{seviri} L1B using Satpy (\texttt{reader=seviri\_l1b\_native}), then match each \textsc{mtg} band to its closest \textsc{msg} counterpart by spectral region and physical quantity (reflectance vs brightness temperature). We restrict to the overlapping subset of channels for which a clear correspondence exists. We then resample one scene to the other's native \texttt{area} (default: \texttt{MSG\_TO\_MTG}) using Satpy's \texttt{Scene.resample} with a configurable resampler (bilinear in our generation pipeline). Optionally, we crop both scenes to a lat/lon bounding box (\texttt{crop(ll\_bbox=...)}). After resampling, arrays share the same spatial grid and shape, enabling direct extraction of aligned channel-wise LR/HR patches.

\subsection{Metric Definitions}
We evaluate super-resolution quality using three complementary metrics.
\textbf{Peak Signal-to-Noise Ratio (PSNR)} measures pixel-level reconstruction 
fidelity between the super-resolved output $\hat{y}$ and the high-resolution 
reference $y$:
\begin{equation}
    \text{PSNR} = 10 \cdot \log_{10}\!\left(\frac{\text{MAX}^2}{\text{MSE}(\hat{y},\, y)}\right),
\end{equation}
where $\text{MAX}$ is the maximum pixel intensity and $\text{MSE}$ is the mean 
squared error. Higher values indicate closer pixel-level agreement.
\textbf{Gradient Preservation Ratio (Grad.)} measures how faithfully the model 
recovers physically meaningful fine-scale cloud structure, defined as the ratio 
of mean gradient magnitudes in the super-resolved output versus the HR reference:
\begin{equation}
    \text{Grad.} = \frac{\mathbb{E}[\|\nabla \hat{y}\|]}{\mathbb{E}[\|\nabla y\|]},
\end{equation}
where $\nabla$ denotes image gradients computed via a Sobel operator. A ratio 
of $1.0$ is ideal: values below $1$ indicate over-smoothing, while values above 
$1$ indicate over-sharpening or spurious texture.
\textbf{Perceptual Distance Ratio (Percep.)} quantifies perceptual similarity of 
the super-resolved output relative to a bicubic baseline using deep feature 
distances. Let $d(A, B) = \|\phi(A) - \phi(B)\|_2$ denote the LPIPS distance 
between images $A$ and $B$, where $\phi(\cdot)$ extracts VGG feature embeddings. 
The ratio is defined as:
\begin{equation}
    \text{Percep.} = \frac{d(\hat{y},\, y)}{d(\hat{y}_{\text{bic}},\, y)},
\end{equation}
where $\hat{y}_{\text{bic}}$ is the bicubic upsampled output. Values below $1$ 
indicate better perceptual fidelity than the bicubic baseline; lower is better.

\subsection{Additional Qualitative Results}
\begin{figure}[ht]
    \centering
    \includegraphics[width=0.9\linewidth]{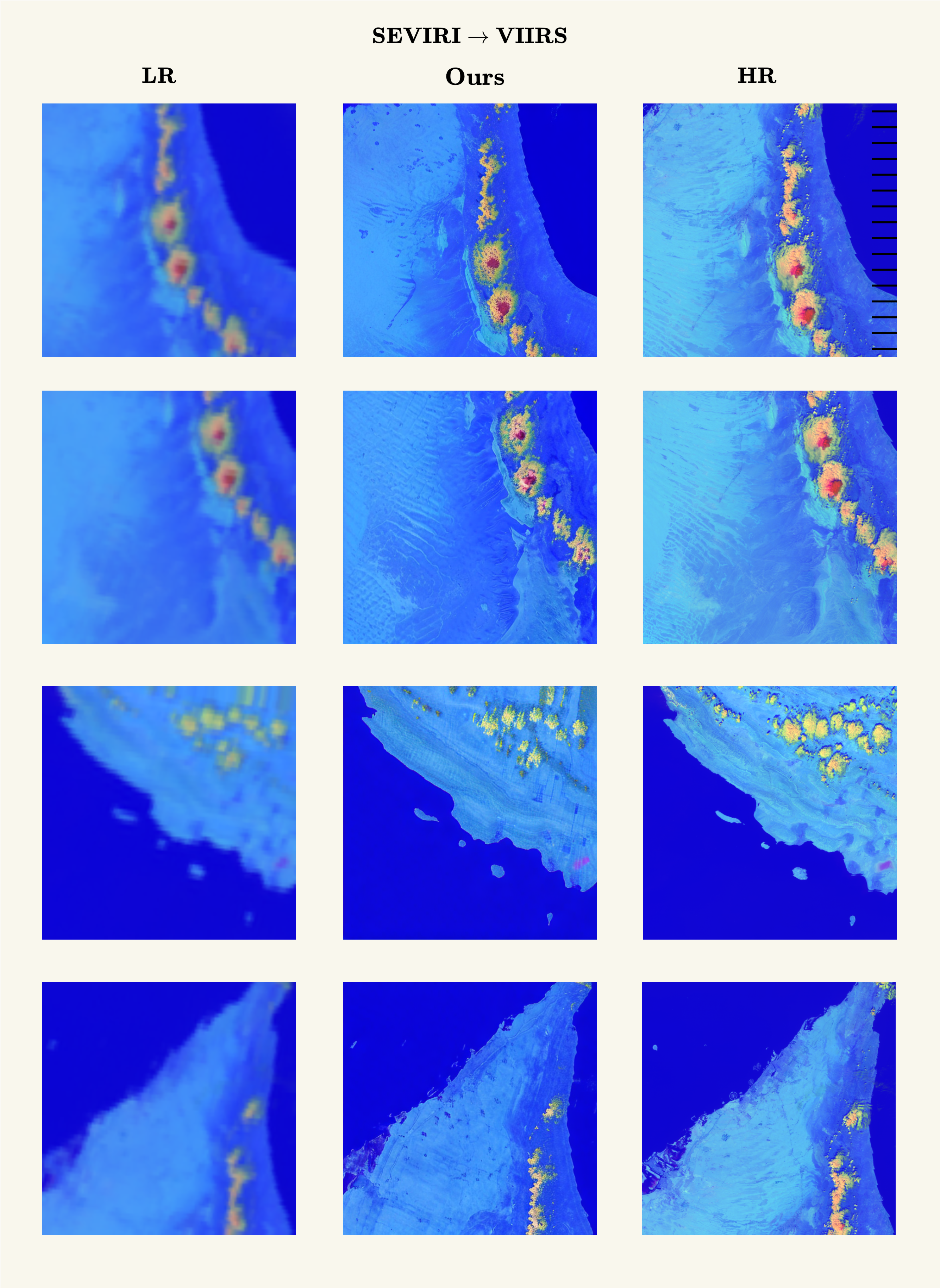}
    \caption{\textbf{Qualitative Comparison \textbf{\textsc{seviri}}$\rightarrow$\textbf{\textsc{viirs}}} (best viewed in zoom).}
    \label{fig:supp-qual-sev-viirs}
    \vspace{-2.5mm}
\end{figure}

\begin{figure}[ht]
    \centering
    \includegraphics[width=0.9\linewidth]{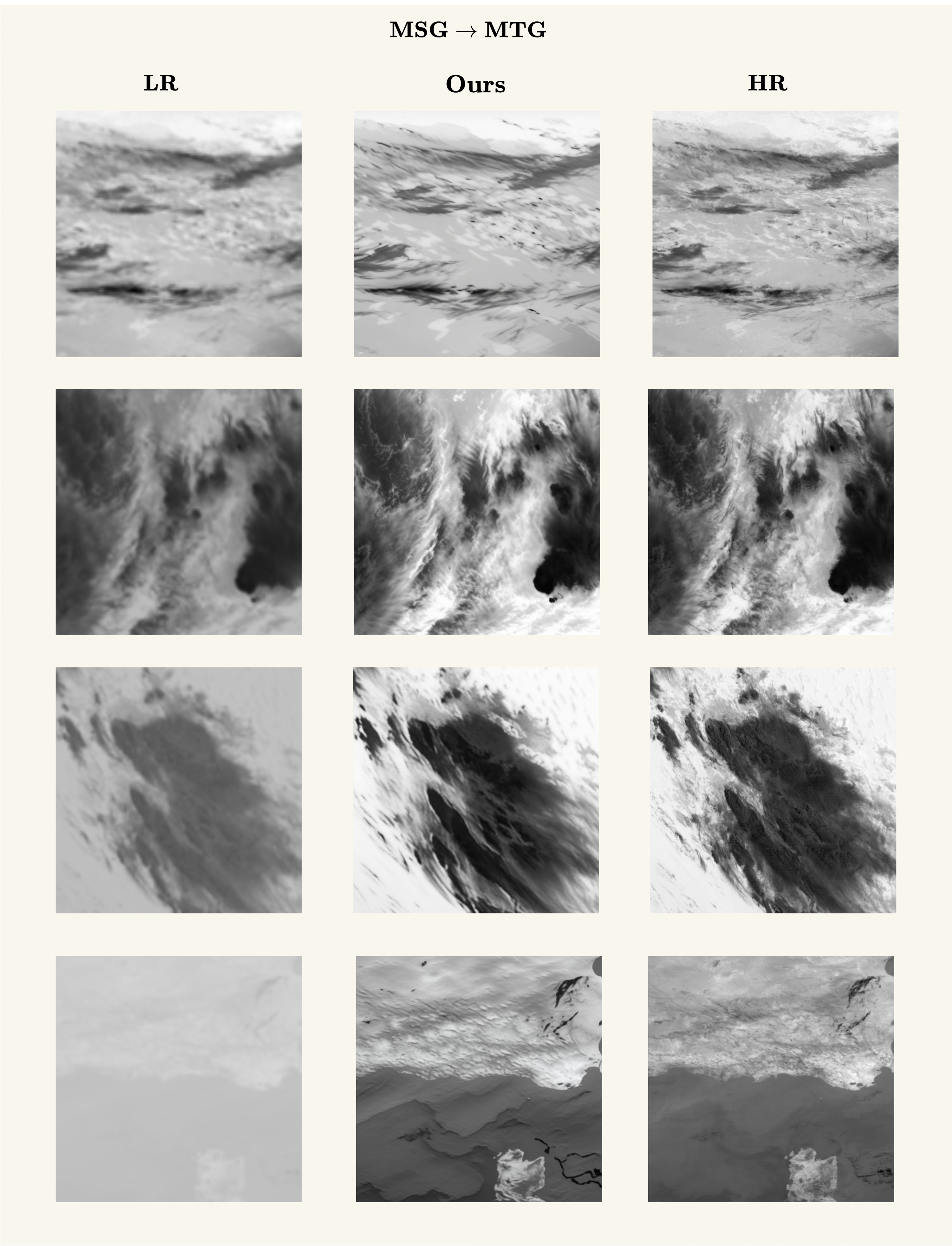}
    \caption{\textbf{Qualitative Comparison \textbf{\textsc{msg}}$\rightarrow$\textbf{\textsc{mtg}}} (best viewed in zoom).}
    \label{fig:supp-qual-msg-mtg}
    \vspace{-2.5mm}
\end{figure}

Figures~\ref{fig:supp-qual-sev-viirs} and \ref{fig:supp-qual-msg-mtg} present further visual comparisons across representative 
cloud scenes for both SEVIRI$\to$VIIRS and MSG$\to$MTG settings, respectively. Across both 
configurations, our method consistently recovers coherent high-frequency cloud 
structures including convective turrets, thin filaments, and sharp cloud-gap 
boundaries. Our 
approach produces outputs that most closely match the ground-truth HR structure 
in both spatial organisation and local gradient fidelity, with less hallucinating 
spurious textures.

\subsection{Reproducibility Statement}
Our code is publicly available at 
\url{https://github.com/hananshafi/superresolution-cloud-microphysics}.
While the UAEREP dataset is subject to institutional data-sharing agreements 
and cannot be released in full, we will release our trained model checkpoints 
and evaluation scripts upon publication to enable usage on publicly available data samples.

\end{document}